\documentclass[runningheads]{llncs}
\usepackage{graphicx}
\usepackage{hyperref}
\usepackage{url}
\usepackage{graphicx}
\usepackage{tikz}
\usepackage{comment}
\usepackage{amsmath,amssymb} 
\usepackage{color}
\usepackage{siunitx}

\begin{document}

\title{Multi Agent Reinforcement Learning of 3D Furniture Layout Simulation in Indoor Graphics Scenes}

%

\author{Xinhan Di\inst{*1}\and Pengqian Yu \inst{*2}}
\institute{Technique Center Ihome Corporation, Nanjing, China\\
\email{deepearthgo@gmail.com}\\
\and
IBM Research, Singapore\\
\email{peng.qian.yu@ibm.com}}

%

\maketitle
\begin{abstract}

In the industrial interior design process, professional designers plan the furniture layout achieve a satisfactory 3D design for selling. In this paper, we explore the interior graphics scenes design task as a Markov decision process (MDP) in 3D simulation, which is solved by multi agent reinforcement learning. The goal is to produce furniture layout in the 3D simulation of the indoor graphics scenes. In particular, we firstly transform the 3D interior graphic scenes into two 2D simulated scenes. We then design the simulated environment and apply two reinforcement learning agents to learn the optimal 3D layout for the MDP formulation in a cooperative way. We conduct our experiments on a large-scale real-world interior layout dataset that contains industrial designs from professional designers. Our numerical results demonstrate that the proposed model yields higher-quality layouts as compared with the state-of-art model. The developed simulator and codes are available at \url{https://github.com/CODE-SUBMIT/simulator2}.
\end{abstract}

\section{Introduction}
Indoors such as the bedroom, living room, office, and gym are important in people life. Function, beauty, cost, and comfort are the keys to the redecoration of indoor scenes. Many online virtual interior tools are developed to help people design indoor spaces in the graphics simulation. 

\begin{figure*}
\centering
\includegraphics[height=6.0cm]{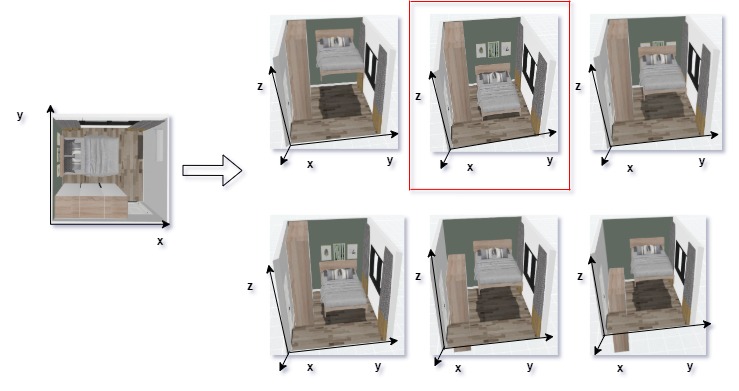}
\caption{An Example of furniture layout in 2D simulation graphic scenes is not practical for producing a good layout for real 3D graphic scenes. Only one of the six 3D layouts(in red block) is good from the same layout solutions in the 2D simulation.}
\label{fig1}
\end{figure*}

Machine learning researchers began to make use of virtual tools to train data-hungry models for the auto layout \cite{Dai_2018_CVPR,Gordon_2018_CVPR}, including a variety of generative models \cite{10.1145/2366145.2366154,10.1145/3303766,Qi_2018_CVPR,10.1145/3197517.3201362}. However, this family of models only provides 2D furniture layout, which is not practical in the real world industry as illustrated in Figure \ref{fig1}.

The prior work neglects the fact that the industrial interior design process in the simulated graphics scenes is indeed a sequential decision-making process in 3D world, where professional designers need to make multiple decisions. This industrial process can be naturally modelled as a Markov decision process (MDP) for 3D simulation. 

The past decade has witnessed the tremendous success of deep reinforcement learning (RL) in the fields of gaming, robotics and recommendation systems \cite{gibney2016google,schrittwieser2020mastering,silver2017mastering}. Researchers have proposed many useful and practical algorithms such as DQN \cite{mnih2013playing} that learns an optimal policy for discrete action space, DDPG \cite{lillicrap2015continuous} and PPO \cite{schulman2017proximal} that train an agent for continuous action space, and A3C \cite{mnih2016asynchronous} designed for a large-scale computer cluster. These proposed algorithms solve stumbling blocks in the application of deep RL in the real world.   
We highlight our two main contributions. First, we develop an indoor graphics scenes simulator and formulate this task as a Markov decision process (MDP) problem. Specifically, we define the key elements of a MDP including state, action, and reward function for the problem. Second, we apply deep reinforcement learning technique to solve the MDP in the learning of the simulated graphic scenes.

\section{Related Work}
Our work is related to data-hungry methods for synthesizing indoor graphics scenes simulations through the layout of furniture. Early work in the scene modeling implemented kernels and graph walks to retrieve objects from a database \cite{Choi_2013_CVPR,Dasgupta_2016_CVPR}. The graphical models are employed to model the compatibility between furniture and input sketches of scenes \cite{10.1145/2461912.2461968}. Besides, an image-based CNN network is proposed to encoded top-down views of input scenes\cite{10.1145/3197517.3201362}. A variational auto-encoder is applied to learn similar simulations \cite{10.1145/3381866,DBLP:journals/corr/abs-1901-06767,Jyothi_2019_ICCV}. Furthermore, the family for the simulation of indoor graphic scenes in the form of tree-structured scene graphs is studied \cite{10.1145/3303766,10.1145/3197517.3201362,10.1145/3306346.3322941}. However, this family of models hard produces accurate size and position for the furniture layout. 

\section{Problem Formulation}
We formulate the process of furniture layout in the 3D simulation of graphics indoor scenes as a Markov Decision Process (MDP) augmented with a goal state $G$ that we would like two agents to learn. We define this MDP as a tuple $(S,G,A,T,\gamma)$, in which $S$ is the set of states, $G$ is the goal, $A$ is the set of actions, $T$ is the transition probability function. 

\begin{figure*}
\centering
\includegraphics[height=7.0cm]{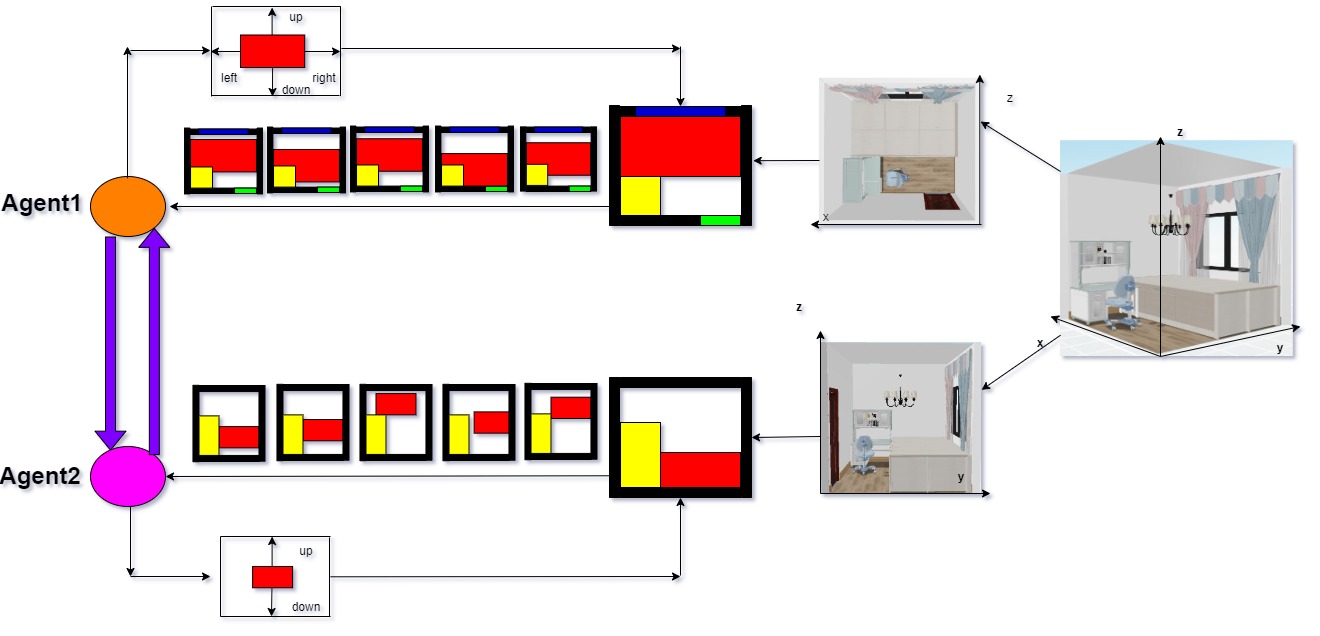}
\caption{The Formulation of MDP for the layout of furniture in the 3D indoor scenes. The 3D graphic scenes simulation is firstly transformed into 2 2D simulation scenes. Two simulators are developed and then deep reinforcement learning is applied to train two agents for the exploration of action, reward, state and optimal policy in a cooperative setting.}
\label{fig2}
\end{figure*}

As shown in Figure \ref{fig2}, the global state $S$ is the geometrical representation of walls, windows, doors and furniture in the 3D simulation graphics scenes including size $s=(x_s,y_s,z_s)$, position $p=(x_p,y_p,z_p)$ of the elements. The global 3D state $S$ is then transformed into two 2D states. $S^{1}$ is the x-y surface of $S$,$S^{2}$ is the y-z surface of $S$. The corresponding actions are $A^{1}$ and $A^{2}$ representing the motion of how the furniture moves to the correct position in each state. Similarly, the corresponding goals are $G^{1}$ and $G^{2}$ representing the correct position of furniture in each state.    

At the beginning of each episode in a MDP at each state $s^{i},i\in\{1,2\}$, the solution to a MDP is a control policy $\pi^{i}: S^{i},G^{i} \rightarrow A^{i}$ that maximizes the value function $v_{\pi^{i}}(s^{i},g^{i}):=\mathbb{E}_{\pi^{i}}[\sum_{t=0}^{\infty} \gamma{i}^{t} R^{i}_{t}|s^{i}_{0}=s^{i},g^{i}=G^{i}]$ for given initial state $s^{i}_0$ and goal $g^{i}$.

\section{Learning Simulations of 3D Indoor Graphics Scenes in a Cooperative Setting}
In order to solve this formulated MDP problem in the 3D simulation of graphics indoor scenes. We explore and develop 2 simulated 2D environment, action, reward, agent and the learning of two cooperative agents in this section as Figure \ref{fig2} shown.

\begin{figure*}
\centering
\includegraphics[height=9.0cm]{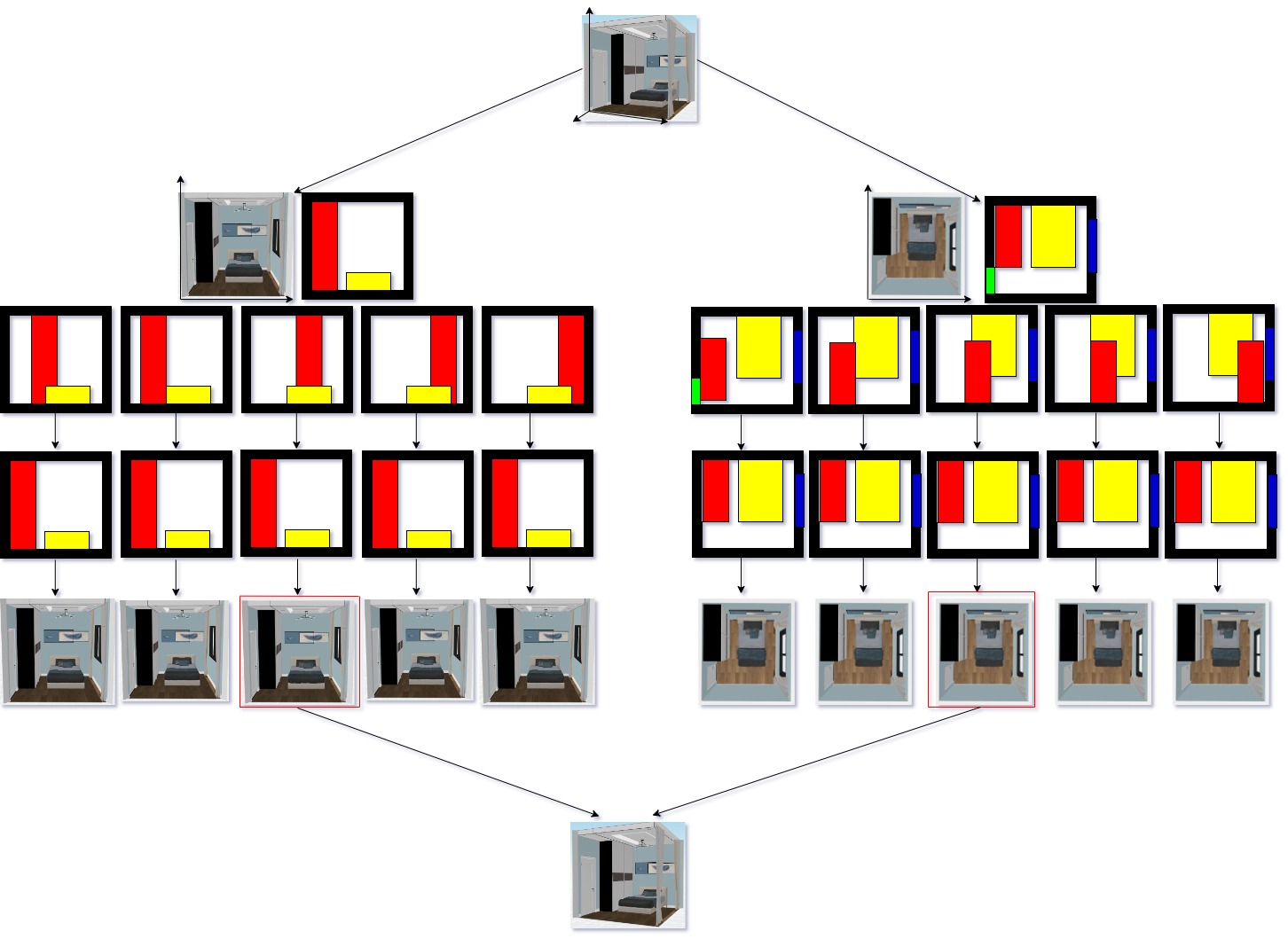}
\caption{The 3D graphic scene(first row) is firstly transformed into 2 2D scenes. Then 2 simulators transfer the real indoor scenes into 2 2D simulated graphics indoor scenes(second row). Two agents produce 2D furniture layout(third row) given initially random position(fourth row) in learning each 2D simulation. Then a good 3D layout is produced in the 3D graphic scene.}
\label{fig3}
\end{figure*}

\subsection{Simulation Environment}
This indoor simulation environment for this defined MDP is implemented as two simulators $F^{i},i \in \{1,2\}$, where $(S^{i}_{next},R^{i})=F^{i}(S^{i},A^{i})$, $A^{i}$ is the action from the agent $i$ in the state $S^{i}$. The simulator $F^{i}$ receives the action $A^{i}$ and produces the next state $S^{i}_{next}$ and reward $R^{i}$.In the next state, the geometrical position, size of walls, doors, and windows are not changed, the geometrical position of the furniture is updated following the input action.

The action $A^{1}$ space is discrete, they are defined as four different action named as right, left, below and up. It means the center of the furniture moves right, left, below and up in a step. Besides, we set another two rules particularly for this simulation. Firstly, in the training process, if the furniture moves beyond the room, the simulator drops this action and then receives a new action. In the test process, if the furniture moves beyond the room, the simulator stops the action immediately. Similarly, the action $A^{2}$ is defined as two different action named as up and down. As Figure \ref{fig2} shown.

We define reward $R^{i}, i \in \{1,2\}$ for the corresponding environment. The reward function works to encourage the furniture to move towards the most right position. It's defined as the following:
\begin{equation}
    R^{i}=\theta_{1}IoU(f_{target},f_{state}), i \in \{1,2\}
\end{equation}
where $\theta_{1}$ is the parameter which is a positive parameter, $f_{target}$ represents the ground truth position and size of the furniture, $f_{state}$ represents the current position and size of the furniture in the state. IoU represents the intersection between $f_{state}$ and $f_{target}$.

DQN \cite{mnih2013playing} algorithm is applied for the learning of each agent in the simulation. It learns an optimal control policy $\pi^{i}: S^{i},G^{i} \rightarrow A^{i}, i \in \{1,2\}$. 

\subsection{Cooperative Learning}
We explore a cooperative learning between agent $1$ and agent $2$ in the learning of 3D simulation graphic scenes. In each iteration of the learning, firstly, agent $1$ moves the furniture in the x-y surface of the 3D graphics scenes, secondly, the next state $s_{t+1}^{1}$ and the intermediate state $s_{t}^{2}$ updates after action $a_{i}^{t}$, thirdly, agent $2$ moves the furniture in the y-z surface up/down. The next state $s_{t+1}^{2}$ updates after $a_{2}^{t}$. Besides, the reward for agent $2$: $R^{2}_{u}$ is updated as following:

\begin{equation}
    R^{2}_{u} = \theta_{2} R^{1} + \theta_{3} R^{2}
\end{equation}

where $\theta_{2}$ and $\theta_{3}$ are positive parameters.

\section{Evaluation}
Both qualitative and quantitative results demonstrating the utility of our proposed model are represented. Four main types of indoor rooms are evaluated including the bedroom, the tatami room, the balcony room and  the kitchen. For each room, we also test the performance of the proposed model in the developed environment with $2000$ random starting points. For the comparison, we train $5000$ samples for each type of rooms and test $1000$ samples for the corresponding type of rooms.We use IoU which is the measure of the intersection between the predicted layout and the ground truth layout in Table \ref{table1}. A two-alternative forced-choice (2AFC) perceptual study is conducted to compare the images from generated scenes with the corresponding scenes from the sold industrial solutions in Table \ref{table2}. The generated 3D layout scenes are generated from our models. To be noted, to the best of our views, the current state-of-the-art models learn simulation in 2D interior graphics scenes. Therefore, comparison between the proposed models and the state-of-the-art models PlanIT \cite{10.1145/3306346.3322941}, LayoutGAN \cite{DBLP:journals/corr/abs-1901-06767} for the $x-y$ surface of the 3D graphic scenes are made. In the perceptual study, the z value is randomly set as PlanIT \cite{10.1145/3306346.3322941}  and LayoutGAN \cite{DBLP:journals/corr/abs-1901-06767} are not for 3D interior graphic scenes. To be noted, we do not compare with layoutVAE  \cite{JyothiDHSM19} and NDN \cite {2020-Lee-NDNGLGWC} since they generates outputs in a conditional manner.

\begin{table}[t]
\centering
\begin{tabular}{|p{2cm}|p{2cm}|p{2cm}|p{2cm}|p{2cm}|p{2cm}}
\hline
\multicolumn{1}{|c|}{}&\multicolumn{4}{|c|}{\text{IoU}}\\
\hline
\hfil Model    &\hfil PlanIT(x-y) &\hfil LayG(x-y)  & \hfil Ours(x-y) & \hfil Ours(y-z)\\
\hline
\hfil tatami    &$0.604\pm0.005$ &$0.645\pm0.013$ &$0.741\pm 0.018$ &$0.738\pm 0.025$\\
\hfil bedroom   &$0.647\pm0.009$ &$0.648\pm0.017$ &$0.765\pm 0.024$ &$0.741\pm 0.031$\\
\hfil balcony   &$0.638\pm0.009$ &$0.629\pm0.019$ &$0.758\pm 0.025$ &$0.735\pm 0.019$\\
\hfil kitchen   &$0.621\pm0.008$ &$0.612\pm0.015$ &$0.734\pm 0.052$ &$0.729\pm 0.027$\\
\hline
\end{tabular}
\caption{Comparison with the state-of-art Model.}
\end{table}\label{table1}

\begin{table*}
		\caption{Percentage ($\pm$ standard error) of 2AFC perceptual study where the real sold solutions(3D) are judged as more plausible than the generated scenes.}
		\begin{center}
			\begin{tabular}{lccc}
				\hline
				Room&Ours(3D)&PlanIT(3D)&LayoutGAN(3D)\\
                tatami&$65.21\pm5.81$&$89.89\pm4.12$&$88.41\pm5.31$\\
                bedroom&$68.36\pm3.96$&$85.27\pm5.49$&$89.75\pm4.62$\\
                balcony&$67.59\pm4.17$&$87.12\pm3.62$&$82.91\pm4.38$\\
                kitchen&$64.29\pm3.59$&$88.73\pm6.32$&$83.59\pm5.61$\\
			\end{tabular}
			\label{table2}
		\end{center}
	\end{table*}

\section{Discussion}
In the learning of 3D simulated interior graphic scenes, we initially formulate a MDP task and learn the simulation of furniture layout through multi agent reinforcement learning in a cooperative way. There are many challenges such as multiple furniture layout, competitive learning of 3D simulation and etc waiting to be explored. 
\bibliographystyle{splncs04}
\bibliography{egbib}

\end{document}